# Enhancement of Healthcare Data Performance Metrics using Neural Network Machine Learning Algorithms

Qi An, Patryk Szewczyk, Michael N Johnstone, James Jin Kang
*Computing and Security, School of Science*
*Edith Cowan University*
Joondalup, WA Australia
{angela.an, p.szewczyk, m.johnstone, james.kang}@ecu.edu.au

*Abstract*— **Patients are often encouraged to make use of wearable devices for remote collection and monitoring of health data. This adoption of wearables results in a significant increase in the volume of data collected and transmitted. The battery life of the devices is then quickly diminished due to the high processing requirements of the device. Given the importance attached to medical data, it is imperative that all transmitted data adhere to strict integrity and availability requirements. Reducing the volume of healthcare data for network transmission may improve sensor battery life without compromising accuracy. There is a trade-off between efficiency and accuracy which can be controlled by adjusting the sampling and transmission rates. This paper demonstrates that machine learning can be used to analyse complex health data metrics such as accuracy and efficiency of data transmission to overcome the trade-off problem. The study uses time series nonlinear autoregressive neural network algorithms to enhance both metrics by taking fewer samples to transmit. The algorithms were tested with a standard heart rate dataset to compare their accuracy and efficiency. The result showed that the Levenberg-Marquardt algorithm was the best performer with an efficiency of 3.33 and accuracy of 79.17%, which is similar to other algorithms' accuracies but demonstrates improved efficiency. This proves that machine learning can improve without sacrificing a metric over the other compared to the existing methods with high efficiency.**

## I. Introduction

Machine learning is a sub-field of artificial intelligence, that has been used in finance, entertainment, spacecraft, pattern recognition, computer version, computational biology, and medical applications [1]. The increased adoption of machine learning applications in healthcare provides medical experts with advanced capabilities to diagnose and treat diseases [2]. Machine learning can support the extraction of relevant data from many patients' datasets stored in electronic health records [3]. Using related disease-causing features from electronic health records, machine learning algorithms can help detect diseases by evaluating data and predicting the underlying causes of the illness [4]. Machine learning provides remarkable advantages for the evaluation and assimilation of the amount of complex health data. Furthermore, machine learning can be deployed for data classification, prediction and, clustering compared to the conventional biostatistical approach [5].

Recently, the rise of smart cities [6], with their mixture of networks, protocols and devices has forced changes to traditional processes in the healthcare and medical industry. For example, mobile health (mHealth) is a prevalent technology using cloud computing, deep learning, artificial intelligence, big data, and machine learning. Health data are collected from patients' wearable sensor devices and forwarded to the hospital database through technologies supporting cellular networks, and then transmitted to cloud storage systems. The collected medical data is then used for further analysis for medical purposes [7]. Some researchers have used machine learning for disease detection, pattern recognition, and image processing from gathered medical data [1-5], [7-8]. Very few studies have provided evidence on machine learning algorithms to improve healthcare data accuracy and efficiency on the network. Many studies [9-13], have focused on improving the trade-off ratio between data accuracy and efficiency using a multilayer inference algorithm. For example, when data accuracy is improved by using a larger quantity of samples, efficiency suffers. Therefore, this paper investigates and evaluates the possibility and feasibility of machine learning method to enhance the healthcare data metrics. Three machine learning time series neural network algorithms including Levenberg-Marquardt, Bayesian Regularization and Scaled Conjugate Gradient are selected to find out which algorithm can enhance healthcare data metrics whilst minimizing the trade-off between accuracy and efficiency.

## II. Methodology

### A. Algorithm selection

Temporal data, also called time series data, exists in many domains such as medicine, video, and robotics. When managing those types of data, a specific method is needed to consider the samples with high correlation between consecutive samples in time series [14]. Artificial neural network (ANN) is a computational intelligence model that can perform time series data classification and regression tasks. It optimizes self-learning to predict the output independent of the provided input [15]. The advantage of neural networks is that it has a better performance for regression task. [16]. This study used heart rate datasets over a period of 1 hour 49 minutes with 6312 data points [17]. Heart rates are predicted based on the historical data pattern with a period sequence. Thus, time series neural network algorithms are appropriate for the experiment.

Three non-linear autoregressive neural network algorithms (Levenberg-Marquardt, Bayesian Regularization, and Scaled Conjugate Gradient) were selected to to implement this study. Levenberg-Marquardt is the most widely used and most effective nonlinear least squares algorithm for neural network training. Its optimization speed is relatively fast, and it has the advantages of fast convergence speed. It is suitable for simple fitting functions or relatively simple parameters to be estimated. In addition, it mostly guaranteed the algorithm for the accuracy for the function approximation type of problems [18]. Bayesian Regularization can quickly minimize errors, and it is suitable for small data sets [19]. Scaled Conjugate Gradient is a fast and efficient training method for neural networks as there is no requirement to set the parameters [20]. MATLAB was used to perform three neural network autoregressive time series - Levenberg-Marquardt, Bayesian Regularization and Scaled Conjugate Gradient. Hence, the collected heart rates were applied to each algorithm to conduct this experiment.

### B. Data split method

Various sampling and training cases were set up to conduct the experiment as shown in TABLE I and tested each scenario with three algorithms to compare and find out the best performer in terms of prediction accuracy and efficiency. The conventional machine learning data division method is 70%-80% data for training, and the rest is for testing. This study split data into 30% to 90% training data to build the different models and seek the possibility of achieving relatively high accuracy and efficiency to reduce the data size to transmit. It is customary to split the data into training, validation and testing data, respectively. The training data presents the neural network training process, while the validation dataset shows the measurement of generalisation and stopping training. The purpose of validation is to avoid overfitting problems. The testing data is to test the performance of neural network algorithms. It provides an independent measurement of network performance.

TABLE I. SAMPLING SCENARIOS OF TRAINING DATASET

|  | Data split method | | |
| --- | --- | --- | --- |
|  | Training data | Validation data | Testing data |
| Scenario 1 | 90% | 5% | 5% |
| Scenario 2 | 80% | 10% | 10% |
| Scenario 3 | 70% | 15% | 15% |
| Scenario 4 | 60% | 20% | 20% |
| Scenario 5 | 50% | 25% | 25% |
| Scenario 6 | 40% | 30% | 30% |
| Scenario 7 | 30% | 35% | 35% |

### C. Data evaluation metrics and software

The machine learning regression technique is deployed to perform the study. MATLAB R2020b is used for applying machine learning algorithms for the training and prediction of heart rate data. The neural net time series deep learning Toolbox 14.1 is selected to generate Levenberg-Marquardt, Bayesian Regulation, and Scaled Conjugate Gradient algorithms.

MSE (mean squared error) and R value from the models evaluate and measure the performance of neural network algorithms. In addition, MAE (mean absolute error) and MAPE (mean absolute percentage error) are calculated after model generation. MAE and MAPE are used to evaluate accuracy (100 – MAPE value). The efficiency can be calculated by (total target data points/trained data points). The 7 scenarios comparison and best-performing scenario for each algorithm are described in the experiment and analysis section.

Formulas were used to evaluate and measure the accuracy and efficiency as below.

MSE is known as mean squared error, $n$ presents the total number of data points (6312 heart rates), $Y_i$ shows the observed values of heart rate where $i$ is a certain data point, ie.1.2.3.4…n, and $\widehat{Y_i}$ is the predicted heart rate value. The statistical parameter calculates the average squared of errors of the predicted heart rates and observed original heart rate values. The less error between them, the higher accuracy is.

$$MSE = \frac{1}{n}\sum_{i=1}^{n}(Y_i - \widehat{Y_i})^2$$

MAE is the mean absolute error value, which presents the average absolute error between the predicted heart rate and the observed heart rate. It directly calculates the average of the residuals. The lower MAE is, the higher accuracy will be.

$$MAE = \frac{\sum_{i=1}^{n}|Y_i - \widehat{Y_i}|}{n}$$

MAPE is the sum of the average absolute error percentage typically for time series forecasting accuracy or error measurement. The MAPE is calculated by the percentage of mean between the observed heart rate value, and the predicted heart rate value. The lower MAPE means the higher accuracy.

$$MAPE = \frac{1}{n}\sum_{t=1}^{n}\left|\frac{Y_i - \hat{Y}_i}{Y_i}\right| \times 100$$

R value is the coefficient of determination that presents the fitness of the predictive model. The large R squared value indicates that there is less error, and the model is more fit for the prediction. Therefore, the highest R value of the model was selected for accuracy. *RSS* means the sum of squares of residuals, *TSS* is the total sum of the squares.

$$R^2 = 1 - \frac{RSS}{TSS}$$

Accuracy of prediction heart rate can be calculated by 100% accuracy minus the mean absolute percentage error value in the model.

$$Accuracy = 100\% - MAPE$$

To save the battery of sensor device, efficiency is considered by reducing the volume of data. It can be calculated by the total amount of data points (heart rates) divided by the training data points (heart rate). $n$ is the total number of heart rates (6312), $t$ presents the total number of training heart rates for each scenario.

$$Efficiency = \frac{n}{t}$$

## III. EXPERIMENT AND ANALYSIS

### A. Levenberg-Marquardt algorithm

Levenberg-Marquardt is an iterative algorithm used to solve nonlinear least squares problems algorithm [15]. This algorithm combines two least squares algorithms-gradient descent and the Gauss-Newton method [17]. The gradient descent method reduces the sum of the squared errors by updating the paraments in the steepest descent. The Gauss-Newton method reduces squared errors by assuming the least square function is locally quadratic, and the parameter estimation value is close to the optimal value range [17]. The combination of the two methods can quickly find the optimal value.

TABLE II provides information about 7 different training scenarios results from the Levenberg-Marquardt algorithm. It is evident that the prediction accuracy is very similar in the range of 79% apart from scenario 5. In terms of the least MSE, and MAPE, higher R value and accuracy and efficiency, the scenario 7 outperforms the other 6 scenarios. The detailed analysis of the scenario is discussed in TABLE III.

TABLE II. LEVENBERG-MARQUARDT TESTING DATA RESULTS

| Training scenarios | MSE | R | MAE | MAPE | Accuracy | Efficiency |
|---|---|---|---|---|---|---|
| Scenario 1 | 0.20 | 0.9981 | 1.43 | 20.51% | 79.48% | 1.11 |
| Scenario 2 | 0.23 | 0.9976 | 1.42 | 20.45% | 79.54% | 1.24 |
| Scenario 3 | 0.23 | 0.9975 | 1.42 | 20.47% | 79.53% | 1.42 |
| Scenario 4 | 0.24 | 0.9977 | 1.46 | 20.92% | 79.08% | 1.66 |
| Scenario 5 | 0.25 | 0.9972 | 1.47 | 21.09% | 78.91% | 2 |
| Scenario 6 | 0.23 | 0.9975 | 1.46 | 20.96% | 79.04% | 2.5 |
| Scenario 7 | 0.21 | 0.9977 | 1.45 | 20.83% | 79.17% | 3.33 |

TABLE III shows the best network performance of the Levenberg-Marquardt algorithm from scenario 7. It uses 30% of samples (1894 heart rates) to train the algorithm, 35% (2209 heart rates) for networking performance validation, and 35% for testing (2209 heart rates).

The lower MSE (0.21) for the test set indicates that the Levenberg-Marquardt algorithm was the best performer of the algorithms tested. Training, validation and testing for R-value are relatively close to 1, especially the testing set result (0.9977) shows the highest value in the entire 7 test scenarios. The MAPE (20.83%) represents the percentage average of the total error data predicted in the test. The model has relatively high accuracy and efficiency with 79.17% and 3.33 respectively.

TABLE III. BEST PERFORMANCE OF LEVENBERG-MARQUARDT FROM SCENARIO 7

| Data split method | Training (%) | Training data points | Validation (%) | Validation data points | Testing (%) | Testing data points |
|---|---|---|---|---|---|---|
| Results | 30% | 1894 | 35% | 2209 | 35% | 2209 |
| MSE | 0.24 | | 0.27 | | 0.21 | |
| R | 0.9975 | | 0.9971 | | 0.9977 | |
| MAE | 1.45 | | | | | |
| MAPE | 20.83% | | | | | |
| Accuracy | 79.17% | | | | | |
| Efficiency | 3.33 | | | | | |

Fig.1 depicts the training and validation MSE (0.27) at 17 epochs. Fig. 2 shows the gradient and validation checks result. The gradient decrease in the cost function with the lowest error point from epoch 17 until 23 shows no error increased since then. In addition, the maximum validation checks are increased from epoch 17 to 23, which reaches the default checks at 6.

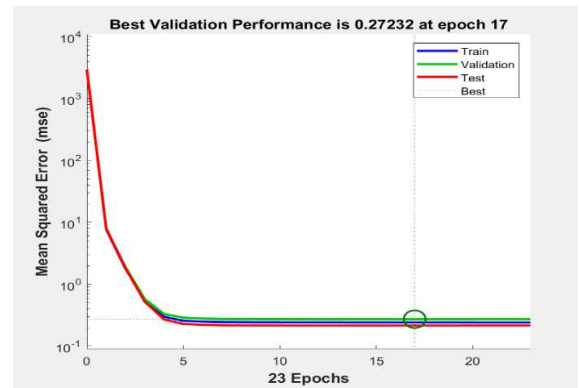

Fig. 1. Performance of Neural Network Performance at epochs 17

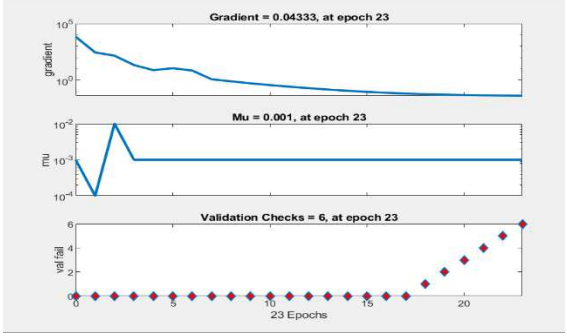

Fig. 2. Training State (Gradient and Validation checks at 6)

Fig. 3 illustrates the histogram of target – output (heart rate prediction) errors with 20 bins. It can be observed that the minimum of prediction errors are allocated with 0.024, which is the closest value to 0 error, with 1700 training data points, 1400 validation data points, and testing is 1800. The fitting tool of the training regression algorithm results can be shown in Fig. 4. The training, validation and testing model R value mentioned in TABLE III are all relatively close to 1 (zero error), especially from the test data (0.9977) with highest value which means almost errorless.

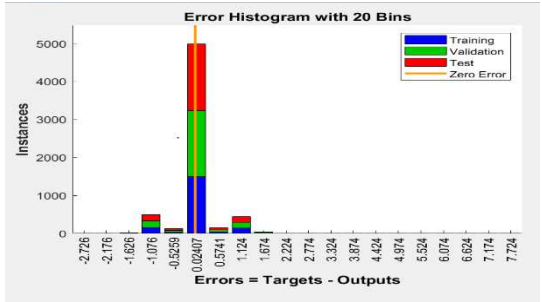

Fig. 3. Error Histogram with lowest error (close to 0 error)

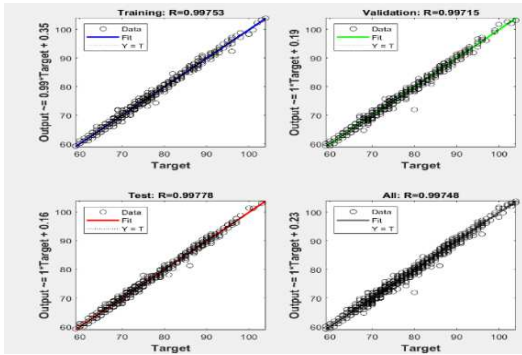

Fig. 4. Training Regression with R (test 0.997)

Fig. 5 shows that the time series reflects seconds of an output (heart rate) and errors. Fig. 6 demonstrates that the autocorrelation lag 0 is equal to the mean squared error, almost 0.25. It displays the same negative and positive values information symmetrically, indicating that, there is a high autocorrelation, and the prediction errors are related the consecutive heart rate. In addition, the correlation of zero lag means the correlation is within 95% of confidence limits, the model is supposed to be adequate for prediction for the accuracy.

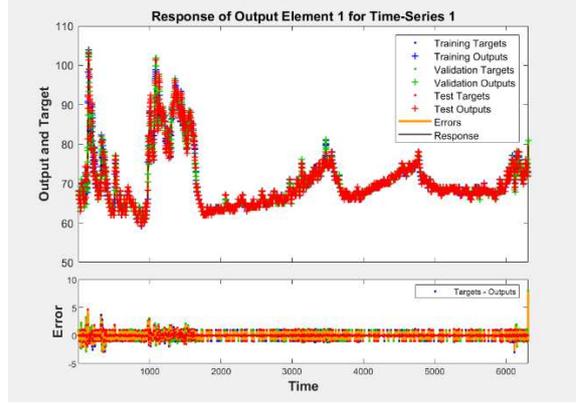

Fig. 5. Output Heart Rate of Time Series Response

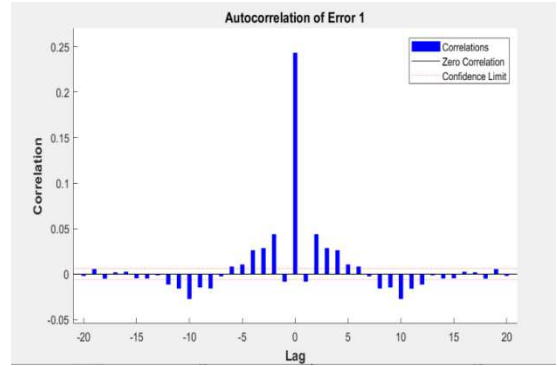

Fig. 6. Autocorrelation of Error

### B. Bayesian Regularization algorithm

Bayesian Regularization is an algorithm using the mathematical process that transforms a nonlinear regression into a "well posed" problem by means of [19] ridge regression, which is used to solve the multicollinearity problem by adjusting the parameter [21]. It is a function of the network that updates the bias and weights values based on the Levenberg-Marquardt algorithm. It minimizes the combination weights and squared errors, then decides the right combination to produce a well generalized network. The Bayesian Regularization is more robust in the regression method, and it is not too overfit and overtrain [19],[22].

TABALE IV shows the testing data set results from 7 scenarios for the Bayesian Regularization algorithm. Scenario 6 has a better performance in connection with lower MSE, higher R value, relatively lower MAPE, higher efficiency, and accuracy compared to the other 6 training scenarios. Hence, scenario 6 was selected for further analysis in TABLE V.

TABLE IV. BAYESIAN REGULARIZATION TESTING DATA RESULTS

| Training scenarios | MSE | R | MAE | MAPE | Accuracy | Efficiency |
|---|---|---|---|---|---|---|
| Scenario 1 | 0.20 | 0.9973 | 1.44 | 20.67% | 79.33% | 1.11 |
| Scenario 2 | 0.32 | 0.9965 | 1.49 | 21.35 | 78.65% | 1.24 |

| Training scenarios | MSE | R | MAE | MAPE | Accuracy | Efficiency |
|---|---|---|---|---|---|---|
| Scenario 3 | 0.28 | 0.9973 | 1.43 | 20.53% | 79.47% | 1.42 |
| Scenario 4 | 0.30 | 0.9970 | 1.44 | 20.64% | 79.36% | 1.66 |
| Scenario 5 | 0.25 | 0.9973 | 1.42 | 20.42% | 79.57% | 2 |
| Scenario 6 | 0.23 | 0.9975 | 1.45 | 20.87% | 79.13% | 2.5 |
| Scenario 7 | 0.29 | 0.9971 | 1.45 | 20.90% | 79.1% | 3.33 |

In scenario 6 in TABLE V, Bayesian Regularization neural network algorithm has the best performance with 40% (2524 heart rate data points) training, 30% (1984 data points) validation, and 30% (1984) testing compared to the other 6 scenarios. The raining and validation sets have the same MSE value (0.24) that indicates the model is stable for prediction. The testing set performs the better result with the lowest error in terms of MSE (0.23). Regarding the R value, the testing set did not achieve the expected result, which we expect to have the highest R value between the 3 data sets. The accuracy (79.13%) of the prediction model is relatively high, with 2.5 of efficiency.

TABLE V. BEST PERFORMANCE OF BAYESIAN REGULARIZATION FROM SCENARIO 6

| Data split method | Training Be (%) | Training data points | Validation (%) | Validation data points | Testing (%) | Testing data points |
|---|---|---|---|---|---|---|
| Results | 40 % | 2542 | 30% | 1984 | 30% | 1984 |
| MSE | 0.24 | | 0.24 | | 0.23 | |
| R | 0.9974 | | 0.9976 | | 0.9975 | |
| MAE | 1.45 | | | | | |
| MAPE | 20.87% | | | | | |
| Accuracy | 79.13% | | | | | |
| Efficiency | 2.5 | | | | | |

Fig. 7 shows the best performance of the Bayesian Regularization algorithm scenario 6 at epoch 23. The algorithm runs 29 epochs, and it can be found that the training and validation are the best performance with MSE value 0.24, 0.23. The model is trained without overfitting, and it is the best performance for the validation. Fig. 8 indicates the gradient and validation checks result for Bayesian Regularization Algorithm scenario 6. The Gradient algorithm is decreased at the lowest error at epoch 29. The validation check reaches the time of 6.

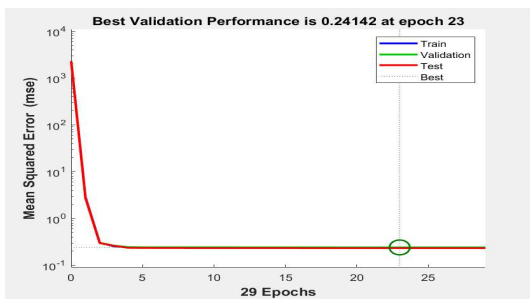

Fig. 7. Performance of Neural Network at epoch 23

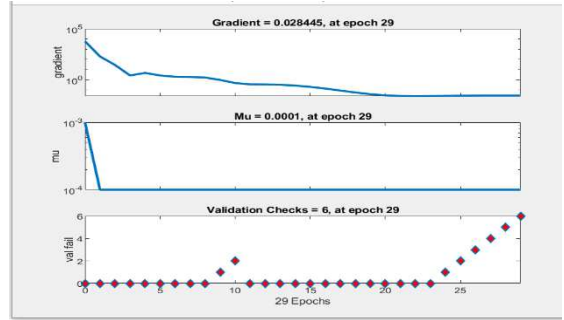

Fig. 8. Training State (Gradient and Validation checks 6)

The histogram in Fig. 9 shows 2000 training data close to zero error, while validation is 1500, and testing is 1300. Fig. 10 reveals the neural network training regression results. It is evidence that at epoch 29, the model is most stable with R value (training 0.9974, validation 0.9976, testing 0.9975). After 29 iterations, the model has the least error.

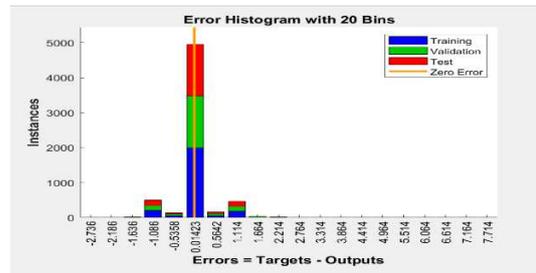

Fig. 9. Error Histogram with the lowest error (close to 0 error)

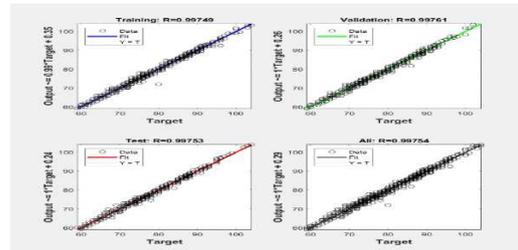

Fig. 10. Training Regression with R (test 0.9975)

Fig. 11 provides the information of time series pattern response of output training errors, validation and testing heart rate data. The autocorrelation plot in Fig. 12, MSE is 0.25 at lag 0, shows that there is a positive relationship between the past heart rate value in prediction of the future heart rate.

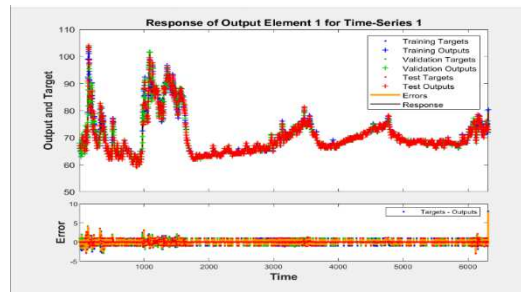

Fig. 11. Output Heart Rate of Time Series Response

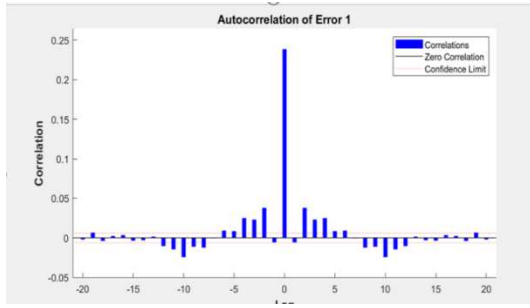

Fig. 12. Autocorrelation of Error

## C. Scaled Conjugate Gradient algorithm

The Scaled Conjugate Gradient algorithm is based on the conjugate direction. Its nonlinear search is different from other conjugate algorithms that require the linear search, and the search is selected in each iteration [18]. Scaled Conjugate Gradient aims to avoid time-consuming linear searches and make the algorithm faster than others [18]. It is a network training function, which can update the weights and bias value according to the scaled conjugate method. From an optimization perspective, neural networks learn to minimize the function, namely a multivariate function that depends on the weight of the network. The algorithm presents quadratic approximation s of the error [18]. As long as its input, weights and transfer function have derivative functions, it can train any network. The step size is the quadratic approximation function of the error function, which makes it more independent and robust for user-defined parameters [19].

In TABLE VI, it lists the testing result from scaled conjugate algorithm 7 training scenarios. Training scenario 5 relatively outperforms other scenarios 6, with low MSE, MAPE, efficiency, and accuracy. The detailed analysis of scenario testing is shown in TABLE VII.

TABLE VI. SCALED CONJUGATE GRADIENT TESTING DATA RESULTS

| Training Scenarios | MSE | R | MAE | MAPE | Accuracy | Efficiency |
|---|---|---|---|---|---|---|
| Scenario 1 | 0.29 | 0.9969 | 1.98 | 28.38% | 71.62% | 1.11 |
| Scenario 2 | 0.24 | 0.9973 | 1.81 | 26% | 74% | 1.24 |
| Scenario 3 | 0.34 | 0.9967 | 1.7 | 24.5% | 75.5% | 1.42 |
| Scenario 4 | 0.37 | 0.9962 | 2.39 | 34.35% | 65.65% | 1.66 |
| Scenario 5 | 0.29 | 0.9966 | 1.89 | 27.02% | 72.98% | 2 |
| Scenario 6 | 0.33 | 0.9967 | 2.02 | 29.07% | 70.93% | 2.5 |
| Scenario 7 | 0.39 | 0.9960 | 1.96 | 28.11% | 71.89% | 3.33 |

TABLE VII. demonstrates the fitted model with Scaled Conjugate Gradient algorithm scenario 5 with 50% training data, 25% validation, and 25% testing data heart rate. Regarding MSE, the model is not performing well as the testing set has the highest MSE (0.29) compared to the training and validation set. The R value for all three data sets is relatively close to 1. Training and validation sets are performed better as the almost same value. However, the testing result for R value ranks the lowest value with 0.9966, which indicates the model is not stable. The model has 72.98% prediction accuracy with a relatively lower efficiency of 2.

TABLE VII. BEST PERFORMANCE OF SCALED CONJUGATE GRADIENT FROM SCENARIO 5

| Data split method | Training (%) | Training data points | Validation (%) | Validation data points | Testing (%) | Testing data points |
|---|---|---|---|---|---|---|
| Results | 50% | 3156 | 25% | 1578 | 25% | 1578 |
| MSE | 0.26 | | 0.28 | | 0.29 | |
| R | 0.9973 | | 0.9972 | | 0.9966 | |
| MAE | 1.89 | | | | | |
| MAPE | 27.02% | | | | | |
| Accuracy | 72.98% | | | | | |
| Efficiency | 2 | | | | | |

Fig. 13 shows the Scaled Conjugate Gradient algorithm performance with MSE declining MSE with increasing epochs, which is an ideal situation. The best validation performance with MSE 0.28 stopped at epoch 67 in the total of epochs 73 to avoid the model overfitting. The Scaled Conjugate Gradient Neural Network Training State in Fig. 14 provides the Gradient and validation checks information. The validation checks are increased at epoch 30, 40, 47, and 63 until reaching time 6 at epoch 73. The is no doubt that the Gradient is locally decreasing with the increasing epochs until 73. However, the decline is not stable, and it fluctuates over a small range.

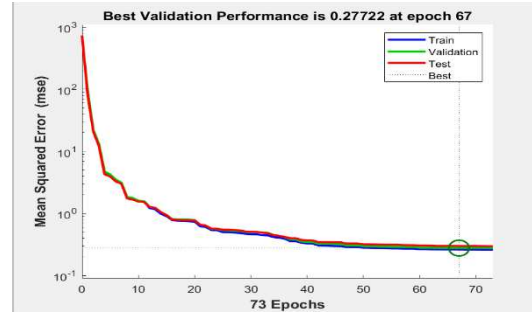

Fig. 13. Performance of Neural Network at epoch 67

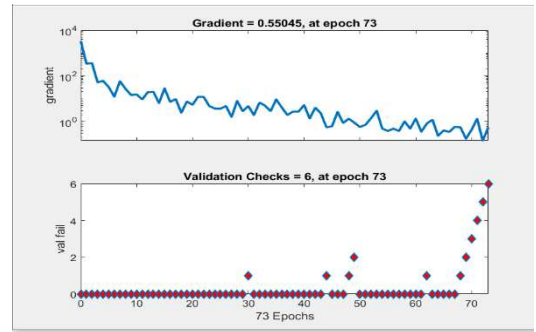

Fig. 14. Training State (Gradient and Validation checks 6)

Fig. 15 indicates the error data distribution with 20 bins in the histogram plot. The error ranges from -1.283 to 1.577. Most of the predicted error data is allocated to the value of -0.7106 for training, validation, and testing. The small amount of error data is close to zero, with 2300 training, 1100 validation, and 1000 testing. The Neural Network training regression algorithm results from Fig. 16 proclaims the fitted regression model with R value at epoch 73 with R value of 0.9971 representing all the data predicted for the algorithm.

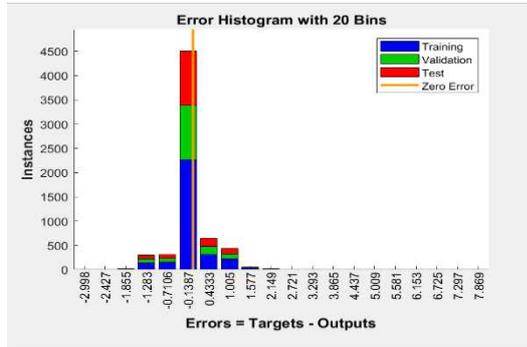

Fig. 15. Error Histogram with the lowest error (close to 0 error)

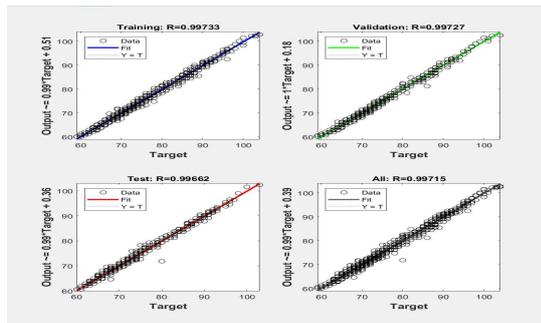

Fig. 16. Training Regression with R (test 0.9966)

Fig. 17 shows time series plot response graph of all information about training data for targets and outputs, validation data for targets and outputs, testing data for targets and outputs and prediction error with time response with the best performance at epochs of 73. In Fig. 18, the autocorrelation plot indicates that the error of MSE is 0.27 at lag 0. As the lag increase, the autocorrelation is relatively low, resulting in a high prediction error.

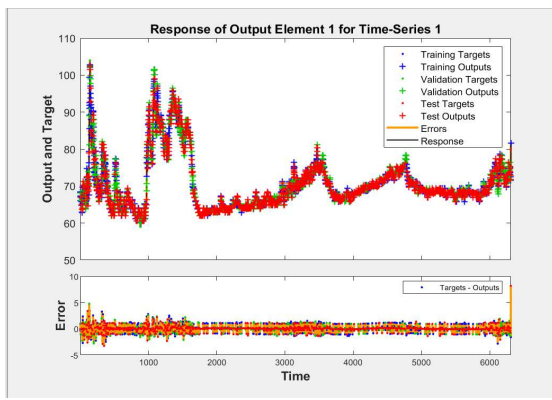

Fig. 17. Output Heart Rate of Time Series Response

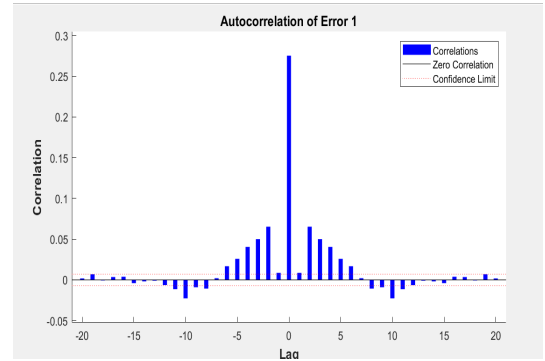

Fig. 18. Autocorrelation of Error

## IV. Discussion

The experiment used three Nonlinear Autoregressive Algorithms across seven different train/validate/test data scenarios and evaluated the best algorithm model in terms of R-value, Mean Squared Error, Mean Absolute Percentage Error, and Accuracy. The experiments in this study showed that the Levenberg-Marquardt neural network algorithm had stable performance in all seven tests. Especially in scenario 7, 30% training data, 35% validation data and 35% testing data are deployed to estimate the neural network performance. TABLE VIII highlights the least error with MSE (0.21), MAE (1.45), and the R (0.9977). Compared with the other Bayesian Regularization and Scaled Conjugate Gradient algorithms, it outperforms with the lowest MAPE (20.83%) and the highest prediction accuracy (79.17%), and efficiency of 3.33.

Regarding scenario 6, Bayesian Regularization neural network performance is inferior to the Levenberg-Marquardt algorithm. It has a higher error in terms of MSE, MAE, MAPE, and lower R value, efficiency, and accuracy. When applying different ratios of training/testing data, the Bayesian Regularization is not stable with respect to training state in gradient and validation checks. The best performance from scenario 6 (40% data for training, 30% for validation, and 33% for testing) provided the least MSE (0.23), MAE (1.45), MAPE (20.87%), R (0.9975), 79.13% accuracy rate, efficiency of 2.5.

The Scaled Conjugate Gradient algorithm scored the worst for the entire 7 scenarios tested regarding the least error evaluation matrix compared to the abovementioned algorithms. Scenario 5 divided data into 50% training, 25% validation and 25% testing to simulate the optimal model. The result shows that the algorithm is not superior to the two other algorithms. It has the highest MSE (0.29), MAE (1.89), MAPE (27.02%), and lowest accuracy of 72.98%, and efficiency of 2.

TABLE VIII. Comparison of 3 Non-Linear Autoregressive Algorithms with Acc (Accuracy), Eff (Efficiency)

| Algorithms | R | MSE | MAE | MAPE | Acc | Eff |
|---|---|---|---|---|---|---|
| Levenberg-Marquardt (Scenario 7) | 0.9977 | 0.21 | 1.45 | 20.83% | 79.17% | 3.33 |
| Bayesian Regularization (Scenario 6) | 0.9975 | 023 | 1.45 | 20.87% | 79.13% | 2.5 |
| Scaled Conjugate Gradient (Scenario 5) | 0.9966 | 0.29 | 1.89 | 27.02% | 72.98% | 2 |

## V. CONCLUSION AND FUTURE WORK

This research applied three algorithms with time series nonlinear autoregressive neural networks (Levenberg-Marquardt, Bayesian Regularization, and Scaled Conjugate Gradient) to predict heart rate. The Levenberg-Marquardt neural network algorithm performance outweighs Bayesian Regularization and Scaled Conjugate Gradient in the matter of the least error and higher prediction accuracy and efficiency. Therefore, machine learning has proved to improve healthcare data metrics simultaneously compared to the existing methods, which trade-off accuracy and efficiency. It is worthwhile to attempt different data ratios with algorithms and compare the model with a high level of accuracy and efficiency. The performance of metrics may vary depending on the size of the datasets for the mature of machine learning algorithms, and the bigger size of the data may result in a better outcome. Whilst the accuracy doesn't change much, the efficiency can improve significantly. This means that it is feasible to improve efficiency by reducing the sample size with maintaining similar accuracy using machine learning algorithms. This study continues to extend the methods with other healthcare types and size to determine the best metrics by optimization of training versus prediction samples.

## REFERENCES

[1] I. El Naqa and M. J. Murphy, "What is machine learning?" in *machine learning in radiation oncology*: Springer, 2015, pp. 3-11. https:// doi: 10.1007/978-3-319-18305-3_1.

[2] U. Sinha, A. Singh, and D. K. Sharma, "Machine learning in the medical industry," in *Handbook of Research on Emerging Trends and Applications of Machine Learning*: IGI Global, 2020, pp. 403-424. https://doi.org/10.4018/978-1-5225-9643-1.ch019.

[3] M. Chen, Y. Hao, K. Hwang, L. Wang, and L. Wang, "Disease prediction by machine learning over big data from healthcare communities," *IEEE Access,* vol. 5, pp. 8869-8879, 2017. https://doi.org/10.1109/ACCESS.2017.2694446.

[4] K. Y. Ngiam and W. Khor, "Big data and machine learning algorithms for healthcare delivery," *The Lancet Oncology,* vol. 20, no. 5, pp. e262-e273, 2019. https://doi.org/10.1016/S1470-2045(19)30149-4.

[5] A. Garg and V. Mago, "Role of machine learning in medical research: A survey," *Computer Science Review,* vol. 40, p. 100370, 2021. https://doi.org/10.1016/j.cosrev.2021.100370.

[6] Z. Baig, P. Szewczyk, C. Valli, P. Rabadia, P. Hannay, M. Chernyshev, M. Johnstone, P. Kerai, A. Ibrahim, K. Sansurooah, N. Syed, and M. Peacock, Future Challenges for Smart Cities: Cyber-Security and Digital Forensics. Digital Investigation, vol 22, 3-13, Aug. 2017. https://doi.org/10.1016/j.diin.2017.06.015.

[7] K. N. Qureshi, S. Din, G. Jeon, and F. Piccialli, "An accurate and dynamic predictive model for a smart M-Health system using machine learning," *Information Sciences*, vol. 538, pp. 486-502, 2020. https://doi.org/10.1016/j.ins.2020.06.025.

[8] K. Shailaja, B. Seetharamulu, and M. Jabbar, "Machine learning in healthcare: A review," in *2018 Second international conference on electronics, communication and aerospace technology (ICECA)*, 2018: IEEE, pp. 910-914. https://doi.org/10.1109/ICECA.2018.8474918.

[9] J. J. W. Kang, "An inference system framework for personal sensor devices in mobile health and internet of things networks," Deakin University, 2017.

[10] J. J. Kang, M. Dibaei, G. Luo, W. Yang and X. Zheng, "A Privacy-Preserving Data Inference Framework for Internet of Health Things Networks," *2020 IEEE 19th International Conference on Trust, Security and Privacy in Computing and Communications (TrustCom)*, 2020, pp. 1209-1214.https://doi: 10.1109/TrustCom50675.2020.00162.

[11] J. J. Kang, "A Military Mobile Network Design: mHealth, IoT and Low Power Wide Area Networks," *2020 30th International Telecommunication Networks and Applications Conference (ITNAC)*, 2020, pp. 1-3. https://.doi: 10.1109/ITNAC50341.2020.9315168.

[12] J. J. Kang, W. Yang, G. Dermody, M. Ghasemian, S. Adibi and P. Haskell-Dowland, "No Soldiers Left Behind: An IoT-Based Low-Power Military Mobile Health System Design," in *IEEE Access*, vol. 8, pp. 201498-201515, 2020. https://.doi: 10.1109/ACCESS.2020.3035812.

[13] J. J. Kang, "Systematic Analysis of Security Implementation for Internet of Health Things in Mobile Health Networks," in *Data Science in Cybersecurity and Cyberthreat Intelligence*, L. F. Sikos and K.-K. R. Choo, Eds. Cham: Springer International Publishing, 2020, pp. 87-113. https:// doi: 10.1007/978-3-030-38788-4_5.

[14] R. Tavenard *et al.*, "Tslearn, A Machine Learning Toolkit for Time Series Data," *J. Mach. Learn. Res.,* vol. 21, no. 118, pp. 1-6, 2020.

[15] P. Kulkarni, V. Watwe, T. Chavan, and S. Kulkarni, "Artificial Neural Networking for remediation of methylene blue dye using Fuller's earth clay," *Current Research in Green and Sustainable Chemistry,* p. 100131, 2021. https://doi.org/10.1016/j.crgsc.2021.100131.

[16] O. I. Abiodun, A. Jantan, A. E. Omolara, K. V. Dada, N. A. Mohamed, and H. Arshad, "State-of-the-art in artificial neural network applications: A survey," *Heliyon,* vol. 4, no. 11, p. e00938, 2018. https://doi.org/10.1016/j.heliyon.2018.e00938.

[17] D. Liu, David, M. Görges, and S. Jenkins, "University of Queensland Vital Signs Dataset: Development of an Accessible Repository of Anesthesia Patient Monitoring Data for Research". *Anesthesia and analgesia*. 114. 584-9. 2011. https://doi.org/10.1213/ANE.0b013e318241f7c0.

[18] H. P. Gavin, "The Levenberg-Marquardt algorithm for nonlinear least squares curve-fitting problems," *Department of Civil and Environmental Engineering, Duke University*, pp. 1-19, 2019.

[19] D. Sardana and S. Srinivasa, "Solar Storm Prediction," 2016.

[20] L. Babani, S. Jadhav, and B. Chaudhari, "Scaled conjugate gradient based adaptive ANN control for SVM-DTC induction motor drive," in *IFIP International Conference on Artificial Intelligence Applications and Innovations*, 2016: Springer, pp. 384-395. https://doi.org/10.1007/978-3-319-44944-9_33.

[21] B. Kibria and S. Banik, "Some ridge regression estimators and their performances," 2020.

[22] D. Selvamuthu, V. Kumar, and A. Mishra, "Indian stock market prediction using artificial neural networks on tick data," *Financial Innovation,* vol. 5, no. 1, pp. 1-12, 2019. 5:16 https://doi.org/10.1186/s40854-019-0131-7.